# A rigorous definition of axial lines: ridges on isovist fields

Rui Carvalho[(1)] and Michael Batty[(2)]

rui.carvalho@ucl.ac.uk    m.batty@ucl.ac.uk

[(1)] The Bartlett School of Graduate Studies
[(2)] Centre for Advanced Spatial Analysis

University College London,
1-19 Torrington Place, London WC1E 6BT, UK

## Abstract

We suggest that 'axial lines' defined by (Hillier and Hanson, 1984) as lines of uninterrupted movement within urban streetscapes or buildings, appear as ridges in isovist fields (Benedikt, 1979). These are formed from the maximum diametric lengths of the individual isovists, sometimes called viewsheds, that make up these fields (Batty and Rana, 2004). We present an image processing technique for the identification of lines from ridges, discuss current strengths and weaknesses of the method, and show how it can be implemented easily and effectively.

## *Introduction: from local to global in urban morphology*

Axial lines are used in space syntax to simplify connections between spaces that make up an urban or architectural morphology. Usually they are defined manually by partitioning the space into the smallest number of largest convex subdivisions and defining these lines as those that link these spaces together. Subsequent analysis of the resulting set of lines (which is called an 'axial map') enables the relative nearness or accessibility of these lines to be computed. These can then form the basis for ranking the relative importance of the underlying spatial subdivisions and associating this with measures of urban intensity, density, or traffic flow. To date, progress has been slow at generating these lines automatically. Lack of agreement on their definition and lack of awareness as to how similar problems have been treated in fields such as pattern recognition, robotics and computer vision have inhibited explorations of the problem and only very recently have there been any attempts to evolve methods for the automated generation of such lines (Batty and Rana, 2004; Ratti, 2001).

One obvious advantage of a rigorous algorithmic definition of axial lines is the potential use of the computer to free humans from the tedious tracing of lines on large urban systems. Perhaps less obvious is the insight that mathematical procedures may bring about urban networks, and their context in the burgeoning body of research into the structure and function of complex networks (Albert and Barabási, 2002; Newman, 2003). Indeed, on one hand urban morphologies display a surprising degree of universality (Batty and Longley, 1994; Carvalho and Penn, 2003; Frankhauser, 1994; Makse et al., 1995; Makse et al., 1998) but little is yet known about the transport and social networks embedded within them (but see (Chowell et al., 2004)). On the other hand, axial maps are a substrate for human navigation and rigorous extraction of axial lines may substantiate the development of models for processes that take place on urban networks which range from issues covering the efficiency of navigation, the way epidemics propagate in cities, and the vulnerability of network nodes and links to failure, attack and related crises. Further, axial maps are discrete models of continuous systems and one would like to understand the consequences of the transition to a discrete approach.

In what follows, we hypothesise a method for an algorithmic definition of axial lines inspired by local properties of space, which eliminates both the need for us to define convex spaces and to trace "(…) all lines that can be linked to other axial lines without repetition" (Hillier and Hanson, 1984, p 99). A definition of axial lines (global entities) with neighbourhood methods (local entities) implies that transition from small to large-scale urban environments carries no new theoretical assumptions and that the computational effort grows linearly (less optimizations) with the number of mesh points used. Our main goal is to gain insight into urban networks in general and axial lines in particular. Therefore we leave algorithm optimizations for future work. It is, however, beyond the scope of the present note to address generalizations of axial maps or to integrate current theories with GIS (but see (Batty and Rana, 2004; Jiang et al., 2000)).

## *The method: Axial lines as ridges on isovist fields*

Axial maps can be regarded as members of a larger family of axial representations (often called skeletons) of 2D images. There is a vast literature on this, originating with the work of Blum on the Medial Axis Transform (MAT) (Blum, 1973; Blum and Nagel, 1978), which operates on the object rather than its boundary (see (Tonder et al., 2002) for a link between Visual Science and the MAT applied to a Japanese Zen Garden). Geometrically, the MAT uses a circular primitive. Objects are described by the collection of maximal discs, ones which fit inside the object but in no other disc inside the object. The object is the logical union of all of its maximal discs. The description is in two parts: the locus of centres, called the symmetric axis and the radius at each point, called the radius function, $R$ (Blum and Nagel, 1978). The MAT employs an analogy to a grassfire. Imagine an object whose border is set on fire. The subsequent internal convergence points of the fire represent the symmetric axis, the time of convergence for unity velocity propagation being the radius function (Blum and Nagel, 1978).

An isovist is the space defined around a point (or centroid) from which an object can move in any direction before the object encounters some obstacle. In space syntax, this space is often regarded as a viewshed and a measure of how far one can move or see is the maximum line of sight through the point at which the isovist is defined. We shall see that the paradigm shift from the set of maximal discs inside the object (as in the MAT) to the maximal straight line that can be fit inside its isovists holds a key to understanding what axial lines are.

As in space syntax, we simplify the problem by eliminating terrain elevation and associate each isovist centroid with a pair of horizontal coordinates $(x, y)$ and a third coordinate - the length of the longest straight line across the isovist at each point which we define on the lattice as $\Delta_{i,j}^{\max}$ where $(x, y)$ is uniquely associated with $(i, j)$. Our hypothesis states that all axial lines are ridges on the surface of $\Delta_{i,j}^{\max}$. The reader can absorb the concept by "embodying" herself in the $\Delta_{i,j}^{\max}$ landscape: movement along the perpendicular direction to an axial line implies a decrease along the $\Delta_{i,j}^{\max}$ surface; and $\Delta_{i,j}^{\max}$ is an invariant, both along the axial line and along the ridge. Our hypothesis goes further to predict that the converse is also true, i.e., that up to an issue of scale, all ridges on the $\Delta_{i,j}^{\max}$ landscape are axial lines. Most of what follows is the development of a method to extract these ridges from the $\Delta_{i,j}^{\max}$ surface, in the same spirit that one would process temperature values sampled spatially with an array of thermometers.

Our method follows a procedure similar to the Medial Axis Transform (MAT). Indeed, the MAT approach to skeletonization first calculates a scalar field for the object (the Distance Map) and then identifies a set of ridge points, or generalized local maxima, in this scalar map. In a discretized representation, the final skeleton consists of such ridge points with the possible addition of a set of points necessary to form a connected structure (Simmons and Séquin, 1998).

Here we sample isovist fields by generating isovists for the set of points on a regular lattice (Batty, 2001; Ratti, 2001; Turner et al., 2001). This procedure is standard practice in spatial modelling (Burrough and McDonnell, 1998). Specifically, we are interested in the isovist field defined by the length of the longest straight line across the isovist at each mesh point, $(i, j)$. This measure is denoted the maximum diametric length, $\Delta_{i,j}^{\max}$ (Batty and Rana, 2004), or the maximum of the sum of the length of the lines of sight in two opposite directions (Ratti, 2001, p 204). To simplify notation, we will prefer the former term.

First, we generate a Digital Elevation Model (DEM) (Burrough and McDonnell, 1998) of the isovist field, where $\Delta_{i,j}^{\max}$ is associated with mesh point $(i, j)$ (Batty, 2001; Ratti, 2001). Next, we use a point algorithm to locate the ridges based on their convexity that is orthogonal to a line with no convexity/concavity (Rana and Morley, 2002) on the DEM. Our algorithm detects ridges by extracting the local maxima of the discrete DEM. Next, we use an image processing transformation (the Hough Transform) on a binary image containing the local maxima points which lets us rank the detected lines in the Hough parameter space. Finally, we invert the Hough transform to find the location of axial lines on the original image.

The Hough transform (HT) was developed in connection with the study of particle tracks through the viewing field of a bubble chamber (the detection scheme was first published as a patent of an electronic apparatus for detecting the tracks of high-energy particles). It was one of the first attempts to automate a visual inspection task previously requiring hundreds of man-hours to execute (Leavers, 1993) and is used in computer vision and pattern recognition for detecting geometric shapes that can be defined by parametric equations. Related applications of the HT include detection of road lane markers (Kamat-Sadekar and Ganesan, 1998; Pomerleau and Jochem, 1996) and determination of urban texture directionality (Habib and Kelley, 2001; Ratti, 2001).

The HT converts a difficult global detection problem in image space into a more easily solved local peak detection problem in parameter space (Illingworth and Kittler, 1988). The basic concept involved in locating lines is point-line duality. In an influential paper, Duda and Hart (Duda and Hart, 1972) suggested that straight lines might be usefully parameterized by the length, $\rho$, and orientation, $\theta$, of the normal vector to the line from the image origin. Imagine that there is a ridge line in image space. The normal vector for each point on this line is defined by $\rho = x\cos\theta + y\sin\theta$ where $\rho$ and $\theta$ are the same for any pair of coordinates $(x,y)$. If we then compute all lines passing through each pair of coordinates on the ridge line in terms of their normal vector, count all the length and orientation parameters $(\rho,\theta)$, and then plot these counts in the parameter space defined by $\rho$ and $\theta$, the position of each straight line (ridge) in image space will be marked as a peak in parameter space. This then enables us to define the locations of ridges in image space simply from examining all possible normal vectors for all possible points. In short, each point $P = (x,y)$ in the image space is mapped into a sinusoidal curve in the $(\rho,\theta)$ space, $\rho = x\cos\theta + y\sin\theta$, and points lying on the same straight line in the image plane correspond to curves through a common point in the parameter plane—see Figure 1. The HT specifies a line as follows. Imagine yourself standing on the image plane at the origin of the coordinates, facing the positive y direction —see Figure 1c). Turn a specified angle, $\theta_L$, to your right, and then walk a specified number of pixels forward, $\rho_L$. Turn through 90° and go forward; you are now walking along the required line in the image.

The process of using the HT to detect lines in an image involves the computation of the HT for the entire image, accumulating evidence in an array for events by a voting (counting) scheme (points in the parameter plane "vote" for the parameters of the lines to which they possibly belong) and searching the accumulator array for peaks which hold information of potential lines present in the input image. The peaks provide only the length of the normal to the line and the angle that the normal makes with the $y$-axis. They do not provide any information regarding the length, position or end points of the line segment in the image plane (Gonzalez and Woods, 1992). Our line detection algorithm starts by extracting the point that has the largest number of votes on parameter space, which corresponds to the line defined by the largest number of collinear local maxima of $\Delta_{i,j}^{\max}$, and proceeds by extracting lines in rank order of the number of their votes on parameter space. One of us has previously proposed (Batty and Rana, 2004) rank-order methods as a rigorous formulation of the procedure originally outlined of "first finding the longest straight line that can be drawn, then the second longest line and so on (…)" (Hillier and Hanson, 1984, p 99).

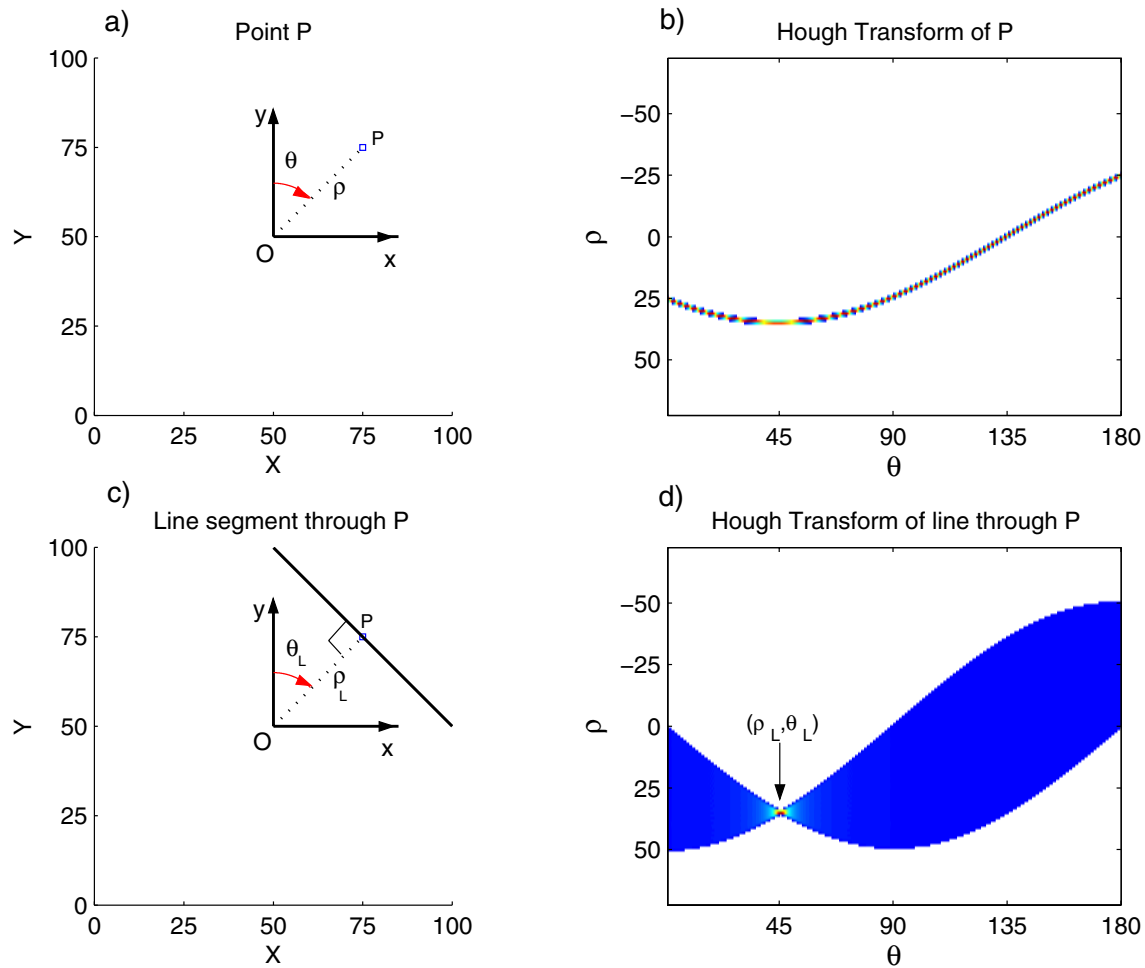

Figure 1. a) Point $P = (75, 75)$ in the image plane. b) The HT converts P into a sinusoidal curve, $\rho = x \cos \theta + y \sin \theta$, where $(x, y) = (25, 25)$ are the coordinates of P relative to $O = (50, 50)$ and $\theta \in [0, 180]$. c) Line segment between points $(0, 50)$ and $(50, 0)$. This segment crosses point P and is orthogonal to the segment $\overline{OP}$. d) The line segment in c) is identified in Hough space by the point where all the sinusoids intersect, $(\rho_L, \theta_L) = (\sqrt{2}x, 45) \cong (35.4, 45)$. The line defined by the segment in c) can be rebuilt on the image plane by starting at $O$ *facing* the direction of the positive $y$ axis, *turning* $\theta_L$ degrees to the right, *walking* forward $\rho_L$ (until $P$) and finally tracing the perpendicular line to $\overline{OP}$.

To test the hypothesis that axial lines are equivalent to ridges on the $\Delta_{i,j}^{\max}$ surface, we start with a simple geometric example: an 'H' shaped open space structure (see Figure 2). As illustrated in Figure 2, axial lines are equivalent to ridges for this simple geometric example, if extended until the borders on the open space.

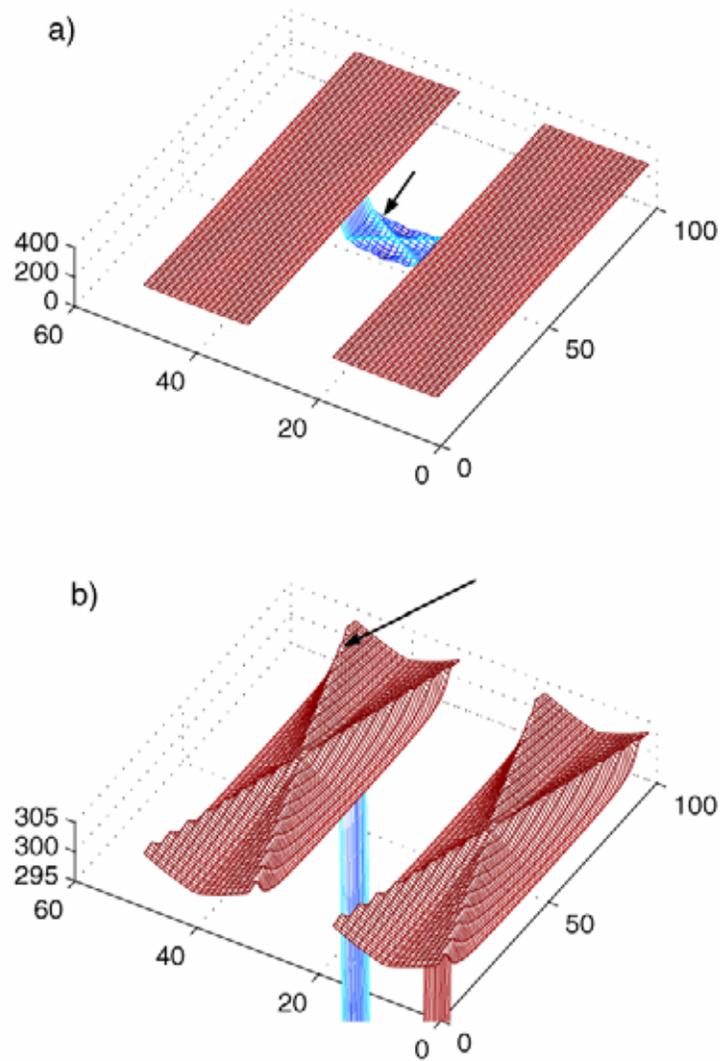

Figure 2. (a) Plot of the Maximum Diametric Length ($\Delta_{i,j}^{\max}$) isovist field for an 'H' shaped open space structure. (b) Zoom-in (detail) of (a) showing the ridges on the longer arms of the 'H' shape. Arrows point to the ridges on both figures.

Indeed, one confirms this both in Figure 2a) and Figure 2b) by properly zooming-in the $\Delta_{i,j}^{\max}$ landscape. Next, we aim at developing a method to extract these ridges as lines by sampling. In Figure 3a), we plot the local maxima of the discretized $\Delta_{i,j}^{\max}$ landscape, which are a discretized signature of the ridges on the $\Delta_{i,j}^{\max}$ continuous field. Figure 3b) is the Hough transform of Figure 3a) where $\theta$ goes from 0° to 180° in increments of 1°. The peaks on Figure 3b) are the maxima in parameter space, $(\rho, \theta)$, which are ranked by height in Figure 3c). The first four visible peaks in parameter space —Figures 3b) and 3c)— correspond to the four symmetric lines defined by the highest number of collinear points in the original space —Figure 3a).

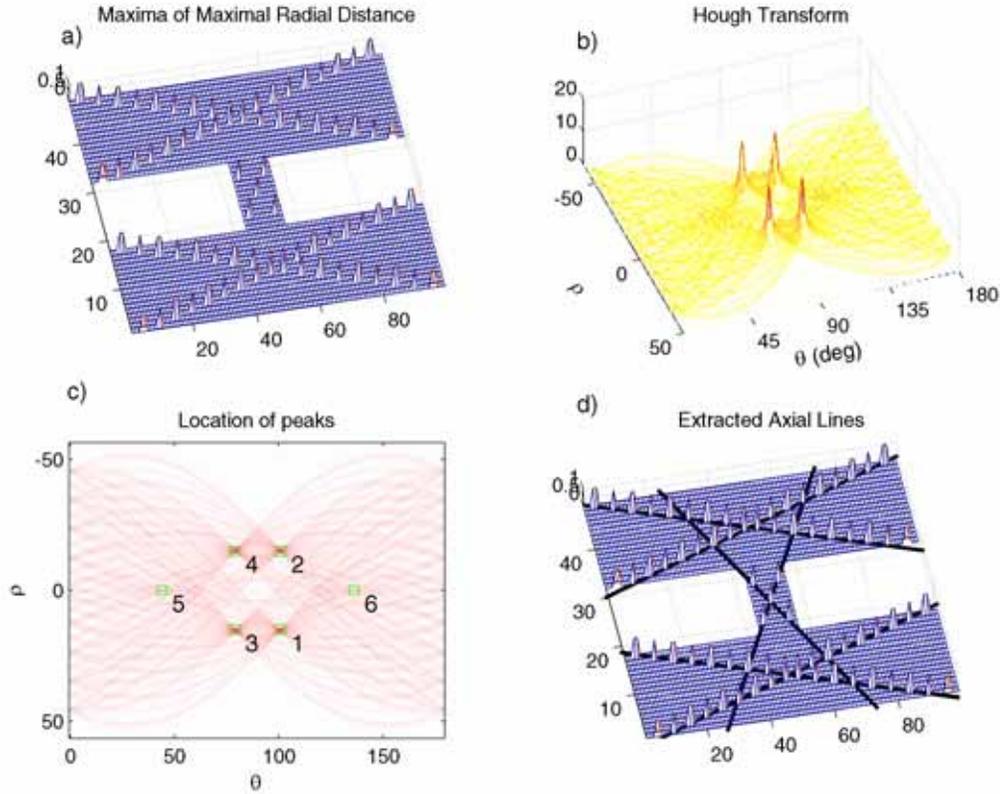

Figure 3. (a) Local maxima of the Maximum Diametric Length ($\Delta_{i,j}^{\max}$) for the 'H' shaped structure in Fig. 1. (b) Hough transform of (a). (c) Rank of the local maxima of the surface in (b). (d) The Hough transform is inverted and the 6 highest peaks in (c) define the axial lines shown.

Finally, the ranked maxima in parameter space are inverted onto the coordinates of the lines in the original space, yielding the detected lines which are plotted on Figure 3d) where we only plot the lines corresponding to the 6 highest peaks in parameter space.

Having tested the hypothesis on a simple geometry, we repeat the procedure for the French town of Gassin —see Figure 4. We have scanned the open space structure of Gassin (Hillier and Hanson, 1984, p 91) as a binary image and reduced the resolution of the scanned image to 300 dpi (see inset of Figure 4).

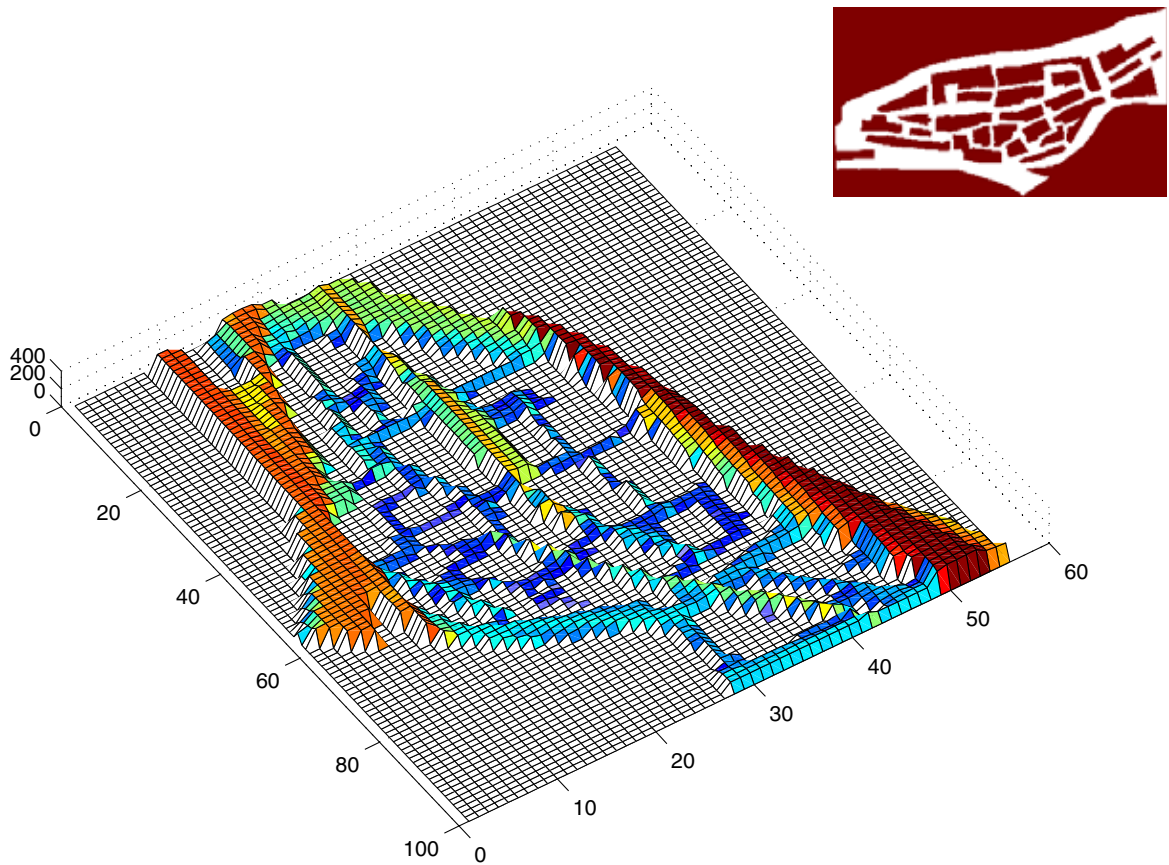

Figure 4. Plot of the Maximum Diametric Length ($\Delta_{i,j}^{\max}$) isovist field for the town of Gassin. The inset shows the scanned image from "The Social Logic of Space" (Hillier and Hanson, 1984).

The resulting image has 171×300 points, and is read into a Matlab matrix. Next we use a ray-tracing algorithm in Matlab (angle step=0.01°) to determine the $\Delta_{i,j}^{\max}$ measure for each point in the mesh that corresponds to open space. The landscape of $\Delta_{i,j}^{max}$ is plot on Figure 4. The next step is to extract the ridges on this landscape. To do this, as we have seen before, we determine the local maxima on the $\Delta_{i,j}^{max}$ landscape. Next, we apply the Hough Transform as in the 'H' shape example and invert it to determine the 6 first axial lines for the town of Gassin (see Figure 5). We should alert readers to the fact that as we have not imposed any boundary conditions on our definition of lines from the Hough Transform, three of these lines intersect building forms illustrating that what the technique is doing is identifying the dominant linear features in image space but ignoring any obstacles which interfere with the continuity of these linear features. We consider that this is a detail that can be addressed in subsequent development of the approach.

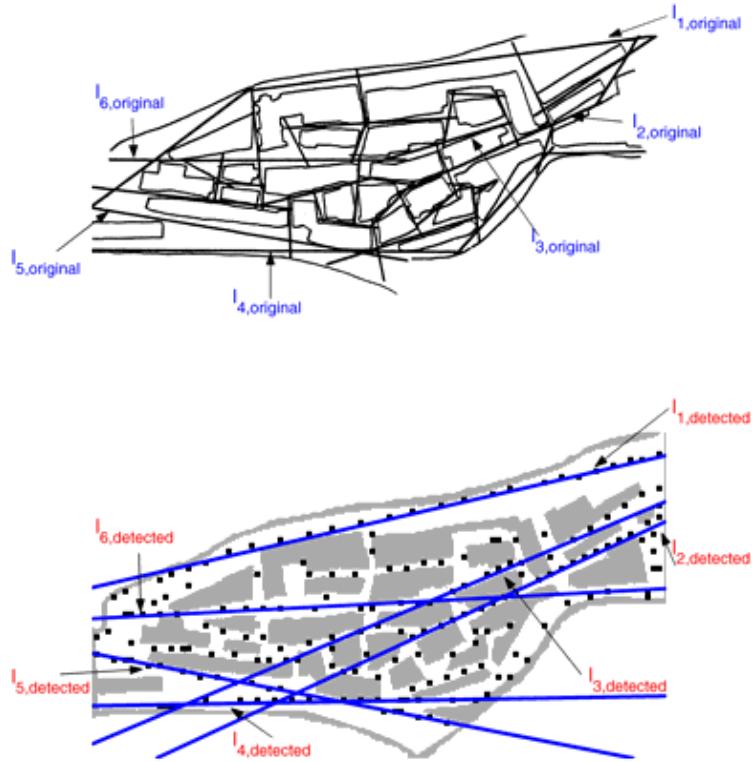

Figure 5. (a) Axial lines for the town of Gassin (Hillier and Hanson, 1984). (b) Local maxima of $\Delta_{i,j}^{max}$ (squares) and lines detected by the proposed algorithm.

## *Discussion: where do we go from here?*

Most axial representations of images aim at a simplified representation of the original image, in graph form and without the loss of morphological information. Therefore, most Axial Shape Graphs are invertible –a characteristic not shared with Axial Maps, as the original shape cannot be uniquely reconstructed from the latter. Also, metric information on the nodes length is often stored together with the nodes (the latter often being weighted by the former), whereas it is discharged in Axial Maps. On the other hand, most skeletonizations aim at a representation of shape as the human observer sees it and therefore aim mostly at small scale shapes (images), whereas the process of generating axial maps assumes that the observer is *immersed* in the shape and aims at the representation of large scale shapes (environments). Nevertheless, we have shown that the extraction of axial lines can be accomplished with methods very similar to those routinely employed in pattern recognition and computer vision (e.g. the Medial Axial Transform and the Hough Transform).

Our hypothesis has successfully passed the test of extracting axial lines both for a simple geometry and for a traditional case study in Space Syntax – the town of Gassin. Indeed, $l_{2,detected}$, $l_{3,detected}$, $l_{4,detected}$, $l_{5,detected}$ and $l_{6,detected}$ in Figure 5 all match reasonably well lines originally drawn (Hillier and Hanson, 1984). Differences between original and detected lines appear for $l_{3,original}$ and $l_{3,detected}$, where the mesh we used to detect lines was not fine enough to account for the detail of the geometry and the HT counts collinear points along a line that intersects buildings, and for $l_{5,original}$ and $l_{5,detected}$, where the original solution is clearly not the longest line through the space.

Figure 5 highlights two fundamental issues which are shared by any spatial problem, both related to the issue of tracing "all lines that can be linked to other axial lines without repetition" (Hillier and Hanson, 1984, p 99). The first is that defining axial lines as the longest lines of sight may lead to unconnected lines on the urban periphery. The problem is quite evident with line $l_{1,original}$ in Figure 5a) (Hillier and Hanson, 1984, p 91), where the solution to the longest line crossing the space is $l_{1,detected}$ —see Figure 5b). This is an expected feature of any spatial problem, in the same way that the existence of solutions to differential equations depends on the given boundary conditions. A possible solution may seem to be to extend the border until lines intersect; nevertheless this may lead both to more intersections than envisioned and disproportionate boundary sizes, as all non-parallel lines will intersect on finite points, but not necessarily near the settlement. Thus, the price to pay for a rigorous algorithm may be that not all expected connections are traced. The second problem is an issue of scale, as one could continue identifying more local ridges with increasing image resolution (see discussion in (Batty and Rana, 2004)). We believe that the problem is solved if the width of the narrowest street is selected as a threshold for the length of axial lines detected from ridges on isovist fields. Only lines with length higher than the threshold are extracted. We speculate that this satisfies almost always the condition that all possible links are established, but are aware that more lines will be extracted automatically than by human-processing (although it does seem that global graph measures will remain largely unaffected by this). Again, this seems to be the price to pay for a rigorous algorithm.

By being purely local, our method gives a solution to the global problem of tracing axial maps in a time proportional to the number of mesh points. This means that algorithm optimization is akin to local optimization (mesh placement and ray-tracing algorithm). Although most of the present effort has been in testing the hypothesis, it is obvious that regular grids are largely redundant. Indeed, much optimization could be accomplished by generating denser grids near points where the derivative of the boundary is away from zero (i.e., curves or turns) to improve detection at the extremities of axial lines. Also, the algorithm could be improved by generating iterative solutions that would increase grid and angle sweep resolutions until a satisfactory solution would be reached or by parallelizing visibility analysis calculations (Mills et al., 1992).

Our approach to axial map extraction is preliminary as the HT detects only line parameters while axial lines are line segments. Nevertheless, there has been considerable research effort put into line segment detection in urban systems, generated mainly by the detection of road lane markers (Kamat-Sadekar and Ganesan, 1998; Pomerleau and Jochem, 1996), and we are confident that further improvements involve only existing theory.

This note shows that global entities in urban morphology can be defined with a purely local approach. We have shown that there is no need to invoke the concept of convex space to define axial lines. By providing rigorous algorithms inspired by work in pattern recognition and computer vision, we have started to uncover problems implicit in the original definition (disconnected lines at boundary, scale issues), but have proposed working solutions to all of them which, we believe will enrich the field of space syntax and engage other disciplines in the effort of gaining insight into urban morphology. Finally, we look with considerable optimism to the automatic extraction of axial lines and axial maps in the near future and believe that for the first time in the history of space syntax, automatic processing of medium to large scale cities may be only a few years away from being implemented on desktop computers.

## *Acknowledgments*

RC acknowledges generous financial support from Grant EPSRC GR/N21376/01 and is grateful to Profs Bill Hillier and Alan Penn for valuable comments. The authors are indebted to Sanjay Rana for using his Isovist Analyst Extension (see http://www.casa.ucl.ac.uk/sanjay/software_isovistanalyst.htm) to provide independent corroboration on the 'H' test problem.